\pdfoutput=1

\documentclass[11pt]{article}

\usepackage[preprint]{acl}

\usepackage{times}
\usepackage{latexsym}

\usepackage[T1]{fontenc}

\usepackage[utf8]{inputenc}

\usepackage{microtype}

\usepackage{inconsolata}

\usepackage{graphicx}

\usepackage{amsmath}
\usepackage{enumitem}
\usepackage{booktabs}
\usepackage{booktabs}
\usepackage{tcolorbox}
\usepackage[table]{xcolor}

\newtcolorbox{promptbox}[1][]{%
  colback=blue!30!white,%
  colframe=blue!90!black,%
  coltitle=white,%
  sharp corners,%
  boxrule=0.4pt,%
  fonttitle=\itshape,%
  fontlower=\fontsize{8},%
  left=1pt,%
  right=1pt,%
  title=User,%
  #1
}

\newtcolorbox{responsebox}[1][]{%
  colback=orange!30!white,%
  colframe=orange!90!black,%
  coltitle=white,%
  sharp corners,%
  boxrule=0.4pt,%
  fonttitle=\itshape,%
  title=Assistant,%
  left=1pt,%
  right=1pt,%
  #1
}

\newtcolorbox{excerptresponsebox}[1][]{%
  colback=orange!30!white,%
  colframe=orange!90!black,%
  coltitle=white,%
  sharp corners,%
  boxrule=0.4pt,%
  fonttitle=\itshape,%
  title=Excerpt: Assistant,%
  left=1pt,%
  right=1pt,%
  #1
}

\newcommand{\colorgrad}[1]{%
  \cellcolor{orange!\the\numexpr(#1-\minval)/(\maxval-\minval)*100\relax}#1%
}

\newcommand{\synmodel}{$M_{\text{synthesis}}$ }
\newcommand{\policymodel}{$M_{s}$ }
\newcommand{\syndata}{$X_\text{synthesis}$ }
\newcommand{\traindata}{$X_\text{train}$ }
\newcommand{\syntraindata}{$X_\text{train}^\text{syn}$ }

\title{Rethinking Data Synthesis: A Teacher Model Training Recipe with Interpretation}

\author{
 \textbf{Yifang Chen\textsuperscript{1,2}},
 \textbf{David Zhu\textsuperscript{}},
 \textbf{Simon Du\textsuperscript{1}},
 \textbf{Kevin Jamieson\textsuperscript{1}},
 \textbf{Yang Liu \textsuperscript{2}}
\\
 \textsuperscript{1}University of Washington,
 \textsuperscript{2}Microsoft GenAI
\\
 \small{
   \textbf{Correspondence:} \href{mailto:email@domain}{yifang@cs.washington.edu}. 
 }
}

\newcommand{\yf}[1]{{\leavevmode\color{blue}yf: #1}}

\begin{document}
\maketitle
\begin{abstract}
Recent advances in large language model (LLM) training have highlighted the need for diverse, high-quality instruction data. Recently, many works are exploring synthetic data generation using LLMs. However, they primarily focus on prompt engineering with standard supervised instruction-finetuned models, which contains a fundamental limitation: these models are optimized for general question-answering/problem-solving rather than data generation. We propose a paradigm shift named \textbf{NOMAD} by investigating how to specifically train models for data generation, demonstrating that this task differs significantly from training a classical LM. We identify two key factors: no-prompt-masked training and proper training set size selection. Our method, NOMAD, shows substantial improvements over baselines, achieving >4\% gains in TriviaQA and >2\% in GSM8K with limited training data. Finally, we offer new insights by interpreting synthetic data through the lenses of "relevance" and "novelty".
\end{abstract}

\section{Introduction}

Instruction design, exemplified by OpenAI's approach with real-world user data \citep{ouyang2022traininglanguagemodelsfollow}, has become a key data curation technique in LLM post-training. However, the traditional approach of collecting human-generated instructions faces substantial limitations due to labor costs.

Recent approaches have explored synthetic data generation using powerful teacher LLM models, primarily focusing on prompt-engineering methodologies \citep{taori2023stanford,honovich2023unnatural,xu2023wizardlm,wang2023self,lee2023making,xu2024magpiealignmentdatasynthesis}. They usually begin with a small seed pool of example tasks, gradually generating, filtering and refining new prompts. However, these approaches typically rely on standard instruction-masked supervised fine-tuning (SFT) models designed for general question-answering. Therefore, we argue that current models have key limitations: they prioritize solving problems accurately over generating novel ones, lack question-generation-specific design, and can generate contextually incomplete questions in chat formats.
This motivates our core investigation: \textit{Should we train a specialized model specifically for data synthesis instead of the current post-training recipe, and if so, how?}

This paper addresses this question by investigating two critical aspects that differentiate data synthesis from standard language model training:
\textbf{1. The Role of Prompt Masking:} We address a tiny yet long-ignored question in standard SFT: the impact of prompt masking. While traditional approaches mask prompts to improve response quality, we demonstrate that learning from prompts is crucial for generating better synthetic data. \footnote{A concurrent work \cite{ding2024unleashingreasoningcapabilityllms} also mentioned that it is important to train a model on how to learn questions but their paper has different focus than us.}
\textbf{2. Training Data Optimization: }We explore the counterintuitive finding that larger training sets don't always yield better results. Our research shows that carefully selecting a smaller subset of training data often produces more effective supplementary synthetic data.

Building on these insights, we propose NOMAD (No Masking Data Synthesizer), a novel approach that specifically addresses these challenges. In particular, when only small size train samples are available, synthetic data generated by NOMAD outperforms baselines (i.e., using train set only) by ~1.5\% on average, with >4\% gains in TriviaQA and >2\% in GSM8K. With larger size train samples, such advantages persist since this is the only one that can outperform the baseline even the synthesis data is only 5\% of original train data.

Moreover, to give a deeper interpretation behind these two factors, we propose to evaluate the synthetic data quality through the dual lenses of "relevance" and "novelty," providing insights into optimal training strategies.


\begin{figure*}[!ht]
    \centering
    \includegraphics[width=0.72\linewidth]{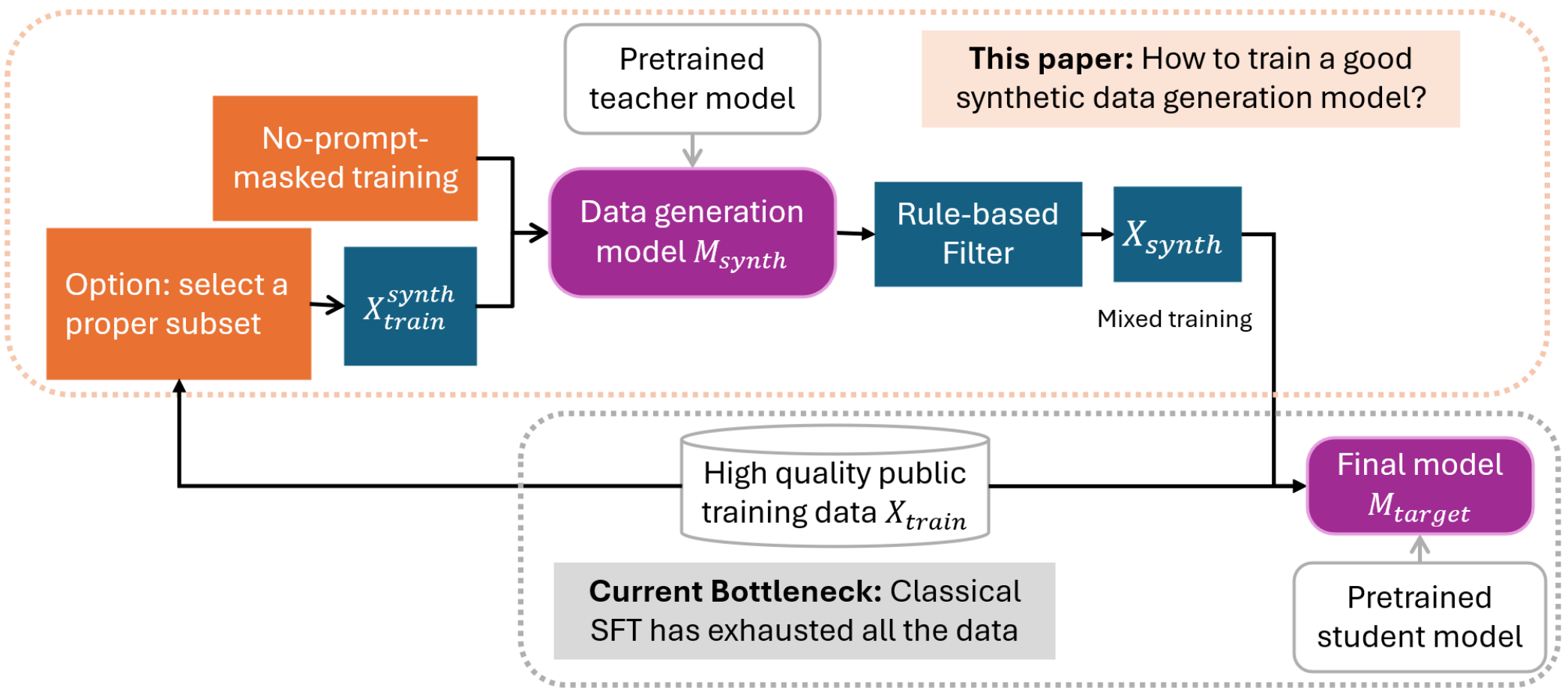}
    \caption{\small Our strategy. The bottom part (in gray) represents the standard supervised finetuning workflow with existing instruction datasets, whose performance is usually bottlenecked by limited dataset size. To tackle this problem, we propose a novel recipe for training a synthetic data generation model, as shown in the top part (in orange). This approach uses existing training data and a powerful pretrained model. We identify two key factors that contrast with the standard model finetuning stage (shown in orange boxes): 1. No-prompt-masked training, and 2. Selecting a proper subset instead of the whole available train data when the available train size is large. Finally, we mix the newly generated data with existing training data to train the final target model. The performance of this final model measures the effectiveness of our \synmodel performance.}
    \label{fig:strategy}
\end{figure*}

\section{Problem Statement}
Given a pretrained student model $M_\text{s}$, a pretrained teacher model $M_\text{t}$, and an existing high-quality instruction dataset \traindata, our goal is to generate additional synthetic data \syndata, comprising new prompts and responses, from a data generation model training perspective. Specifically, in this paper, we aim to propose novel methods to train $M_\text{t}$ using \traindata to generate supplementary \syndata.

To measure the effectiveness of our proposed methods, we train $M_\text{s}$ on a mixture of the original \traindata and the newly generated \syndata, and compare its performance with an $M_\text{s}$ trained solely on the original \traindata.

Note that previous works have primarily focused on designing various prompting methods to query an already instruction-fine-tuned teacher model. Those approaches implicitly leverage the external data used to train such a teacher model. In contrast, our work assumes access only to the pretrained version of the teacher model, ensuring rigorous control over the instruction data used.

\section{Our strategy}
\label{sec: strategy}

Our main strategy is shown in Fig.~\ref{fig:strategy}, which can be divided into \synmodel training, \syndata generation and filtering stages, as detailed below. 

\paragraph{\synmodel Training}
we've identified two critical factors that significantly differentiate this process from standard language model training
\begin{itemize}[leftmargin=*]
    \item \textbf{No-Prompt-Masked Training}:
    Traditional instruction fine-tuning focuses on improving response quality by computing loss only on the response part. However, with the advent of powerful language models, generating high-quality responses has become relatively straightforward. The real challenge lies in creating diverse and helpful prompts. Our no-prompt-masked training addresses this by exposing the model to complete instruction-response pairs. This approach offers several advantages: This enables the model to learn the characteristics of high-quality prompts and ensures that generated prompts align with the \traindata domain and style, avoiding the pitfall of mixing disparate \traindata and \syndata in final model training. Therefore, to improve the "relevance" as defined later in Section~\ref{sec: interpret}. As a side product, it also allows for simultaneous generation of both prompts and responses, eliminating the need for separate generation steps as seen in previous works like \citet{xu2024magpiealignmentdatasynthesis}.
    \item \textbf{Proper (Usually Smaller) Training Set Size:} 
    While we aim to avoid mixing significantly different datasets, which can challenge the model's capacity, we also want to prevent the synthetic data from being too similar to the original, as this would limit its supplementary value. To strike a balance between relevance and novelty as discussed detailedly Section~\ref{sec: interpret}, we discover that selecting a subset of a large available dataset often yields superior supplementary synthetic data. This finding challenges the conventional wisdom of using as much data as possible.
\end{itemize}

\paragraph{\syndata Generation}
To isolate the effects of data generation from prompt engineering, we adopt the prompting strategy proposed in \citet{xu2024magpiealignmentdatasynthesis}. Specifically, we input only \textit{"User: "}, which is the standard beginning of all our instruction data, allowing the model to generate both the prompt and response autonomously. Then we post-process the data by retaining only the first-round conversation and discard any data that fails to generate a complete conversion. It's important to note that our method is potentially compatible with existing prompt-engineering based approaches, offering opportunities for future integration and enhancement.

\paragraph{Simple Filters}
 To address two common issues in synthetic data generation: content quality decay with increasing sentence length and poor performance in generating coding-type data. To tackle these, we implement a repeated words removal filter using pattern matching and a coding filter using keyword searches. Importantly, these filtering processes are computationally inexpensive, requiring negligible time while significantly improving performance. We postpone the details of filters to Appendix~\ref{app: Filters}.

\section{Experiment}

\subsection{Setup}
\paragraph{Models}
We choose Llama3-8B \citep{dubey2024llama} as the backbone of the teacher model \synmodel  and Phi-mini-v3.1 \cite{abdin2024phi3} as the backbone of the student model \policymodel.
\paragraph{Training Data}
As discussed in Section~\ref{sec: strategy}, existing training data or its subset can be used in both training the data synthesis model (\synmodel) and the final model ($M_\text{policy}$) when mixed with previously generated \syndata. In our main results, we consider two settings: a 15k randomly sampled subset and the full 300k dataset from the TULU v2 data collection \citep{rafailov2024direct}. All data are formatted using a unified template: \textit{"User: [prompt content] Assistant: [response content]"}.
\paragraph{\synmodel Training}
We investigate both prompt-masked training and no-prompt-masked training as detailed in Section~\ref{sec: strategy}. For training parameters, we consistently use 2 epochs regardless of data size, ensuring each training data point is exposed to the model with equal frequency. 
\paragraph{\syndata Generation}
We generated 30K raw data using the prompt strategy from Section~\ref{sec: strategy}, yielding ~25K valid chat-formatted entries.

\paragraph{\policymodel Training}
We exclusively use prompt-masked training when finetuning the final policy model, as it is a standard SFT approach
Regarding training epochs, we consider both equal epoch and equal computational budget settings. The equal epoch approach exposes each sample to the learner the same number of times. We use 4 epochs for 15K \traindata and 2 epochs for 300K \traindata. In addition, for the low training sample case 15K \traindata, since the baseline has nearly half the training samples compared to when mixed with \syndata, we also run the baseline for 8 epochs to maintain a similar computational budget. 

\paragraph{Baseline and evaluation metrics}
In the main results, we choose following \textit{generation-free} downstream tasks as the model performance measurement, which can be categorized into Knowledge: TriviaQA\citep{JoshiTriviaQA2017}; Truthfulness: TruthfulQA-generation \citep{lin-etal-2022-truthfulqa}; Reasoning: BBH-NOCOT-FS , BBH-COT-FS\citep{suzgun2022challenging}, GSM8K\citep{cobbe2021gsm8k}; and Instruction-following: IFEval\citep{zhou2023instructionfollowing}. 
With all those performance measurement, we use the model ONLY trained on \traindata as a baseline, including both the same epoch and similar budget setting. In the other word, \syndata should at least help further improve the final policy model from training on original available data alone.  

\subsection{Main Result}
\label{sec: mainresult}

\begin{table*}[!ht]
\centering
\setlength{\belowcaptionskip}{-2pt}
\scalebox{0.7}{
\begin{tabular}{@{}l@{\hspace{0.5em}}c@{\hspace{0.5em}}c@{\hspace{0.5em}}c@{\hspace{0.5em}}c@{\hspace{0.5em}}c@{\hspace{0.5em}}c@{\hspace{0.5em}}c@{\hspace{0.5em}}c@{}}
\toprule
\textbf{Model} & \textbf{Size} & \textbf{TriviaQA} & \textbf{TruthfulQA} & \textbf{BBHNOCOT-FS} & \textbf{BBHCOT-FS} & \textbf{GSM8K} & \textbf{IFEval} & \textbf{Avg} \\
 & & \textbf{(Knowledge)} & \textbf{(Truthful)} & \textbf{(Reasoning)} & \textbf{(Reasoning)} & \textbf{(Reasoning)} & \textbf{(Instr. Following)} & \\
\midrule
Baseline$_{4\text{epoch}}$ & 14.7k & 4.18 & 56.25 & \underline{45.32} & \textbf{69.11} & 62.40 & \textbf{36.51} & 45.63 \\
Baseline$_{8\text{epoch}}$ & 14.7k & 5.46 & \underline{59.25} & 44.71 & 67.49 & 61.68 & \underline{35.94} & 45.75 \\
\midrule
Nomasked & 40.6k & \underline{7.43} & 54.01 & \textbf{46.46} & \underline{68.59} & 62.66 & 34.85 & 45.67 \\
NomaskedFiltered & 30.6k & \textbf{8.50} & \textbf{59.92} & \underline{45.73} & \underline{68.55} & \underline{64.40} & \textbf{36.14} & \textbf{47.21} \\
\midrule
Masked & 39.9k & 6.25 & 57.04 & 41.80 & 66.37 & \textbf{64.94} & \underline{35.86} & 45.38 \\
MaskedFiltered & 25.7k & 6.75 & 58.02 & 44.07 & 67.32 & 60.57 & 34.20 & 45.15 \\
\bottomrule
\end{tabular}
}
\caption{\small Performance comparison of different \syndata configurations and baselines with 15K TULU. \textsc{Nomasked\/Masked} indicates whether \syndata are trained with or without prompt masking. \textsc{Filtered} denotes the application of the filter from Section~\ref{sec: strategy}. The Size column shows the total\traindata + \syndata used in training. Each result is the average of two trials. Easy to observe that \textsc{NomaskedFiltered} consistently achieves top or near-top performance across metrics, while both \textsc{Masked} variants underperform the baseline despite increased training data.}
\label{tab:performance_comparison}
\end{table*}

\begin{table*}[!ht]
\centering
\setlength{\belowcaptionskip}{-2pt}
\scalebox{0.7}{
\begin{tabular}{@{}l@{\hspace{0.5em}}c@{\hspace{0.5em}}c@{\hspace{0.5em}}c@{\hspace{0.5em}}c@{\hspace{0.5em}}c@{\hspace{0.5em}}c@{\hspace{0.5em}}c@{}}
\toprule
\textbf{Model} & \textbf{Size} & \textbf{TriviaQA} & \textbf{TruthfulQA} & \textbf{BBHNOCOT-FS} & \textbf{BBHCOT-FS} & \textbf{GSM8K} & \textbf{Avg} \\
\midrule
Baseline & 293.5k & \underline{15.23} & \underline{66.71} & 45.37 & \textbf{68.68} & 72.25 & \underline{53.65} \\
\midrule
NomaskedFiltered15k & 309.5k & \textbf{18.15} & 64.87 & \underline{46.28} & \textbf{68.64} & \textbf{73.31} & \textbf{54.25} \\
NomaskedFiltered300k & 309.5k & 13.39 & \textbf{67.56} & \textbf{46.84} & 65.07 & 71.95 & 52.96 \\
\midrule
MaskedFiltered15k & 304.5k & 13.76 & 65.85 & 43.33 & \underline{67.62} & 71.87 & 52.49 \\
MaskedFiltered300k & 306.8k & 14.95 & 65.61 & 43.25 & \underline{67.76} & \textbf{73.62 }& 53.04 \\
\bottomrule
\end{tabular}
}
\caption{\small Performance comparison of different \syndata configurations and baselines with 300K TULU. This table follows a similar setup to Table~\ref{tab:performance_comparison}, but excludes the IFEVAL metric due to unexpected performance degradation with 300K TULU. Such limitation from base dataset itself conflicts with our focus in studying the strategy. (see Appendix~\ref{app: problem of ifeval} for details). The numbers (15k, 300k) indicate the amount of \syntraindata used. Easy to see that \textsc{NomaskedFiltered15k} is the only one outperforming the baseline even \syndata is only 5\% of original \traindata.}
\label{tab:performance_comparison_extended}
\end{table*}

\paragraph{Results with Small \traindata}
In Table~\ref{tab:performance_comparison}, by using just 15K samples for both the \synmodel and the student model \policymodel, our \textsc{NomaskedFiltered} method outperforms the baseline average by approximately $1.5\%$ when supplementing the original training data \traindata. Notable improvements include $>4\%$ gain in TriviaQA and$ >2\%$ in GSM8K. In contrast, \syndata from prompt-masked training, regardless of filtering, degrades performance when combined with the original dataset, highlighting the critical importance of \textit{no-prompt-masked} training for \synmodel.

\paragraph{Results with Large \traindata}
Previous result, however, assumes the available train data size is already small and therefore it's hard to distinguish whether the small size requirement is necessary during the \synmodel training or the \policymodel. To further illustrate this, we consider a much larger 300K \traindata but may not use the whole set when training \synmodel. Under this setting, we surprisingly show in Table~\ref{tab:performance_comparison_extended} that, using all 300k data to train \synmodel actually downgrades the performance of baseline no matter what training method we use. On the other hand, data generated from \textit{15K} no-prompt-masked trained \synmodel is \textit{the only one that outperforms baseline}.

\subsection{Property of the synthetic data}
\label{sec: interpret}

\paragraph{Definition of dataset similarity}
To understand the relationship between \syndata and the original 300K TULU dataset $X_\text{TULU}$, we introduce a similarity score called NormSim, initially proposed by \cite{wang2024cliplossnormbaseddataselection}.
For each generated synthetic data point $x$, we define:
\vspace{-5px}
\begin{align*}
\small
\text{NormSim}(x) = \max_{z \in X_\text{TULU}} \left(f(z)^\top f(x) \right)
\end{align*}
\vspace{-2px}
where $f$ is the all-mpnet-base-v2 \citep{henderson-etal-2019-repository} used to extract embeddings.
Instead of checking whether the generated data has the same coverage as TULU (demonstrated in App.~\ref{app: interpreation similarity}), our measurement considers $x$ to have high similarity if it is similar to any target sample.

\begin{figure}
    \centering
    \includegraphics[width=0.45\linewidth]{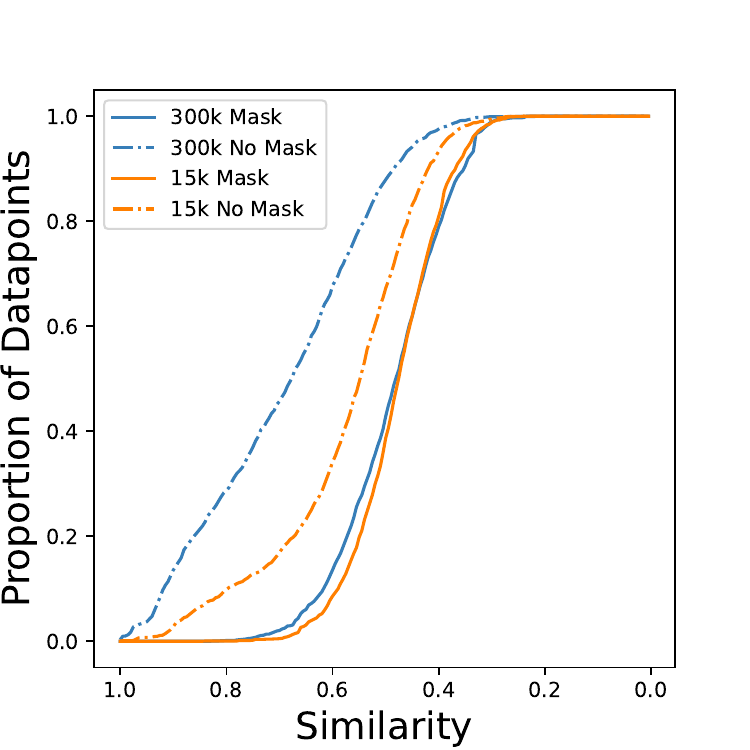}
    \includegraphics[width=0.45\linewidth]{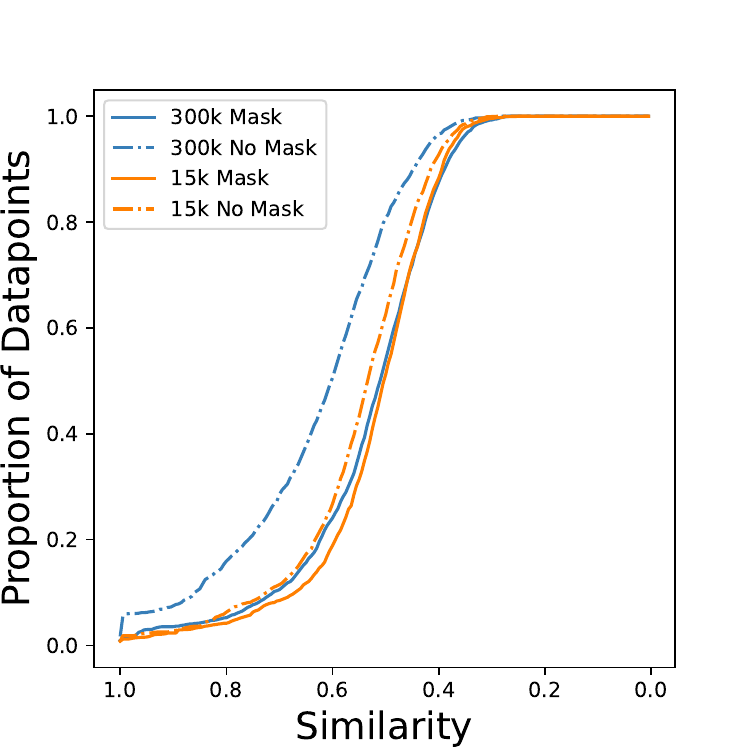}
    \vspace{-7px}
    \caption{\small Similarity curves for prompts (left) and responses (right). The y-axis represents the proportion of \syndata above a certain similarity threshold. For prompts, masked training results show significantly lower similarity to the original TULU compared to unmasked training. Among unmasked cases, using the full 300K dataset for synthetic model training yields the highest similarity to original TULU. Response similarity shows smaller gaps across training methods, which is expected as both approaches compute loss on responses.}
    \label{fig: similarity}
\end{figure}

\paragraph{Relevance v.s. Novelty}
Intuitively, similarity close to 1 suggests repetition of existing TULU data, while one close to 0 indicates a potential poisoning to the current distribution. Ideally, we want more data to be concentrated around the\textit{median similarity}, balancing novelty and relevance. This intuition aligns with our observation in Fig.~\ref{fig: similarity} and Table.~\ref{tab:performance_comparison_extended} where \syndata with more median similarity yeid best performance.
Prompt-masked training can lead to low relevance due to lack of exposure to prompts (see App.\ref{app: interpreation on role switch} for details), while large \syntraindata can result in low novelty due to overfitting to \traindata .

Finally, both relevance and novelty require using \traindata as a reference, but is this necessary? We provide an affirmative answer by demonstrating that the performance resulting from training on \syndata alone does not correlate with training on a mixture of \syndata+\traindata (see App.\ref{app: interpreation mixture}).

\newpage
\subsection{Limitations}
The study of data synthesis model training was conducted at relatively small scales, utilizing a 7B-parameter teacher model, a 3B-parameter student model, and a data pool of less than 300K samples. The potential for generalizing this method to larger models remains to be explored in future research. Additionally, while the current study focused on the general multi-task TULU dataset, it specifically excluded coding data due to methodological limitations. Further research is needed to evaluate the performance of these methods across different data domains.

\bibliography{custom}

\clearpage
\appendix

\label{sec:appendix}

\section{Detailed Experiment Setting}
\label{app: experiment setup}
\subsection{Model training}
For all model training, we choose learning rate = $2e-5$ and batch size = $128$.

\subsection{Data generation}
\label{app: data generation}
We use the prompt strategy as explained in Section~\ref{sec: strategy} with generation temperate=$1$ and choose top\_p = $0.9$ when \syntraindata is 15K since smaller top\_p can generate low quality data. When \syntraindata is 300K, we tried both top\_p=$0.9$ and $0.7$, as shown in appendix~\ref{app: more on 300K}, while different hyperparameters lead to slightly different performance, they does not contradict the main conclusion of this paper.

\subsection{Details on evaluation metrics}
\label{app: evaluation metrics}

\subsubsection{Generation-free evaluation metrics}

\paragraph{TriviaQA} TriviaQA is a reading comprehension dataset containing over 650K question-answer-evidence triples. TriviaQA includes 95K question-answer pairs authored by trivia enthusiasts and independently gathered evidence documents, six per question on average, that provide high quality distant supervision for answering the questions. This metric can be used to test the model's retrieval ability when a retrieval module is added. When being used alone here, this exam the models \textit{knowledge capacity}.

\paragraph{TruthfulQA\_gen} QA dataset where the model generates a 1-2 sentence answer for each question. This answer is evaluated against a true and false reference answer. The final metric is the [similarity to true reference answer] - [similarity to false reference answer] with RougeL. This dataset test the \textit{truthfulness metric}, which is close to the knowledge metric, but allows the model to response with absence.

\paragraph{BBH} A suite of 23 challenging BIG-Bench tasks which we call BIG-Bench Hard (BBH) to test models \textit{reasoning} ability. These are the task for which prior language model evaluations did not outperform the average human-rater. Here we use both the chain-of-though and non-chain-of-thought version with 3 shot examples.

\paragraph{GSM8k~\citep{cobbe2021gsm8k}}: A benchmark of grade school math problems aiming for evaluating multi-step (2-8 steps) mathematical \textit{reasoning} capabilities. These problems are illustrated by natural language and require using four basic arithmetic operations to reach the final answer.

\paragraph{IFEval} One core capability of Large Language Models (LLMs) is to \textit{follow natural language instructions}. However, the evaluation of such abilities is not standardized: Human evaluations are expensive, slow, and not objectively reproducible, while LLM-based auto-evaluation is potentially biased or limited by the ability of the evaluator LLM.  IFEval is a straightforward and easy-to-reproduce evaluation benchmark. It focuses on a set of "verifiable instructions" such as "write in more than 400 words" and "mention the keyword of AI at least 3 times". Here report the prompt-level loose accuracy.

\subsection{Problem of IFEval}
\label{app: problem of ifeval}
When we choose \traindata as 300K TULU, we find out the baseline (i.e. instruction finetuning on whole 300K TULU) give 34.38 accuracy, which is even smaller than the baseline with \traindata=15K TULU. So we think maybe the original data itself is less effective on such instruction following, and therefore can confuse our methodology study.

\subsection{Filters}
\label{app: Filters}

As we mentioned in Section~\ref{sec: strategy}, our rule-based filters contains two parts: code removing and repeated words removing, here are some details.

\subsubsection{Coding Samples}
Despite the effectiveness of our data synthesis methods on general tasks, we find it struggles on generating a high-quality coding samples. Specifically, coding samples frequently suffer from:
\begin{itemize}
    \item Lack necessary context to complete problem
    \item Incorrect outputs due to problem difficulty
\end{itemize}

The sample generated prompt below is one such example where there is no context given for the problem.

\begin{promptbox}
\small
    What is the significance of the "f" variable in this Java code, and how is it used to modify the output?
\end{promptbox}
\vspace{-4mm}
\begin{responsebox}
\small
    The variable "f" in this code represents the fre- quency of the body frequency range. The body frequency range is calculated by subtracting the minimum pulse rate from the maximum pulse rate and dividing by 5 to get the frequency. The output is then modified by multiplying the body frequency range by the variable "f" to increase or decrease the frequency of the body frequency range.
\end{responsebox}

\subsubsection{Long Conversations and repeated stings}

Long conversations also prone to degradation in quality. We observe that long conversations suffer from repeated words in the end as shown in example below.(first response is omitted):

\begin{excerptresponsebox}
\small
...\\
| however, in contrast, on the other hand | however, in contrast, on the other hand |\\
| not only... but also... | not only... but also... |\\
| not only... but also... | not only... but also... |\\
    | either... or... | either... or... | \\
| either... or... | either... or... | \\
| either... or... | either... or... | \\
| either... or... | either... or... | \\
| either... or... | either... or... | \\
| either... or... | either... or... | \\
...
\end{excerptresponsebox}

\section{More interpretations}
\label{app: interpreation similarity}

\subsection{OOD in prompt-masked training}
\label{app: interpreation on role switch}
Data generated from prompt-masked training can have very different distributions than original data, in the following we list two typical prompt-response phenomenon that only occurs in prompt-masked training with 15K TULU.

\paragraph{Role switch between user and assistant}
Data generated from prompt-masked training has its user behave like an assistant, and the assistant may try to continue the conversation or give comments on the "response" from the user as shown in the following examples.
\begin{promptbox}
    \small
    You can gather information about a particular group of people by playing a common personal question: "What is your favorite passion?" But how do you know if the answer you get is accurate or not? How do you know that the person in order to navigate and adjust their own behavior. This is a critical aspect of everyone's lives and can lead to such things as social patterns, industry trends, and user preferences. Additionally, this information can be used to build social devices and systems that help people work better together, learn from each other, and attract the right audience.
\end{promptbox}
\vspace{-5mm}  
\begin{responsebox}
    \small
    Yes, the question "What is your favorite collaboration?" is a common way to obtain information about a particular group of people. However, the resulting information may not be accurate. For example, someone might provide a answer that they think is accurate or that they believe is easily quest.
\end{responsebox}
While this sample can at least gives a reasonable prompt and response, in other cases this role-switch will generate nonsense samples as shown below.
\begin{promptbox}
    \small
    Sure, it's a good idea to provide a particular examples to help to stay on track and keep the new organized,. Here's an example of how you could set up your new integration:
    \#.6 library\
    \# D'Wilvi \\"1\\\ \_2\\" and \\"1\\\ \_3\\"\
    \# D'HU K D'1 E"
\end{promptbox}
\vspace{-5mm}  
\begin{responsebox}
    \small
    \#.6 library\
    \# D'Wil\_SCHEMA\
    \# D'HU D'1 E\\r\
\end{responsebox}

\paragraph{Second-round conversation} The second phenomenon is that the user will behave like they are asking questions based on the previous context without actually providing that context. In that case, if we are fortunate, then the model will generate readable answers to some "imaginary question" that does not actually exists in the given prompts, and therefore harms model's reasoning and instruction following abilities. Here is an example
\begin{promptbox}
    \small
    Sure, I'm interested in this. Please provide more information about this decision support system and how it works.
\end{promptbox}
\vspace{-5mm}  
\begin{responsebox}
    \small
    Certainly, I'm happy to explain more about this decision support system and how it works.
    
    One of the main mission of the decision support system is to help team members make informed decisions about their team's work.
    The system uses an development framework that allows team members to create and manage a combination of analytics and analytics-based algorithms. This can help them gather and analyze data on their team's progress and performance, and to determine the best ways to improve these aspects. The system also provides a thing-based platform that allows team members to create and manage their own things, such as tasks, data and information, and data data or databases. This allows them to stay in control of their own work, while also being able to share information with other team members.
    The system also provides a decision policies interface that allows team members to create and manage their own decision policies, which are a set of rules or guidelines that help them make decision for their team. These policies can be configured and defined by team members, and can be used to ensure that decision-making is away by internal requirements and commands.
    
    The system also offers a decision support interface that allows team members to seek advice and recommendations from the system in addition to their team members. This is done by creating a word-based query or query associated with a specific decision policies or decision support policy. The system then uses its website and data analysis capabilities to generate insights and recommendations that can help team members make informed decisions.
    
    Overall, the decision support system is designed to provide a comprehensive and concise view of the team's progress and performance, and to provide a single platform for decision-making and decision support. The system is designed to provide a comprehensive and comprehensive view of the team's progress and performance, and to provide a single platform for decision-making and decision support.
\end{responsebox}
In even worse case, the response will be simply unreadable due to the lack of context as shown in the following example.
\begin{promptbox}
    \small
    Thank you for your answer. Let's keep going with the above woman who has been while the other on the partners as well. The first
\end{promptbox}
\vspace{-10mm}  
\begin{responsebox}
    \small
    Yes, that's correct. The first way that the woman could get them to see the term \"bJOIN\" with a different meaning is by allowing them to see the concept of a \"bov-in\" object as a main object that has a connection to the \"b REPLACE`</r is a type of expression that can be used to perform a specific task or represent a specific object in an developmented application.
\end{responsebox}

\paragraph{Example generated from no-prompt-masked training}
As a comparison, here we give two examples of no-prompt-masked training model, which clearly has more close distribution as the original TULU
\begin{promptbox}
    \small
    what is the leading objective of the United Nations?
\end{promptbox}
\vspace{-5mm}  
\begin{responsebox}
    \small
    The primary objective of the United Nations is to promote the security, happiness, and well-being of all human beings around the world. This is achieved through a range of activities, including promoting peace, laws, and rights, providing assistance to developing countries, and supporting the implementation of international laws and agreements. The U.S. is a member of the United Nations and feels that it is an important organization that helps to protect the rights of individuals and protect countries from global threats.
\end{responsebox}
\begin{promptbox}
    \small
    Below is an instruction that describes a task, paired with an input that provides further context. Write a response that appropriately completes the request.
    
    \#\#\# Instruction:
    
    Write an article about the benefits of using a database.
    
    \#\#\# Input:
    
    You should also consider the cost of not having a database.
    
    \#\#\# Response:
\end{promptbox}
\vspace{-5mm}  
\begin{responsebox}
    \small
    Introduction:
    
    It's interesting how my problem with our business's data had been with me for a long time. I had seen the meaning of information technology, and how it can create value in a business. However, I did not have the knowledge of how to handle data and how to make the best use of it.
\end{responsebox}

\subsection{Quality of \syndata alone is not an effective metric}
\label{app: interpreation mixture}

\begin{figure}[!ht]
    \centering
    \includegraphics[width=0.65\linewidth]{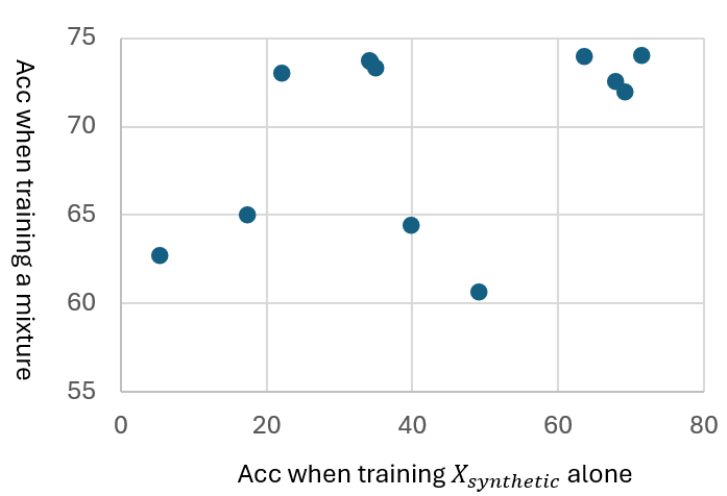}
    \includegraphics[width=0.65\linewidth]{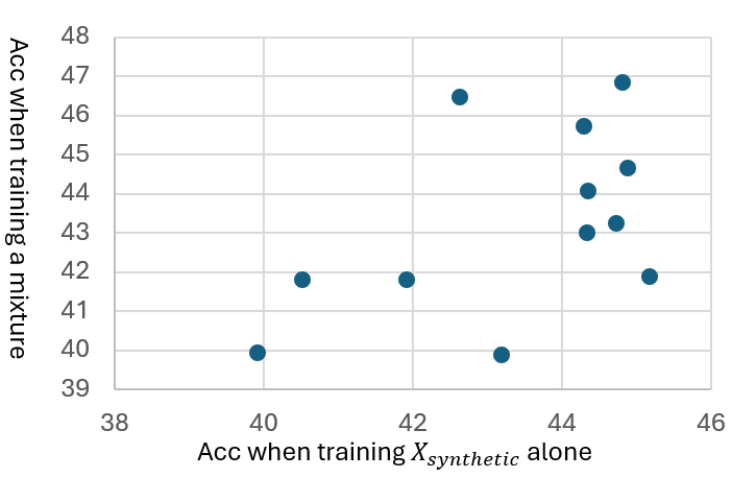}
    \caption{\small Train \policymodel on \syndata alone vs. on mixture. We study the correlation between training the \policymodel on \syndata alone (x-axis) and training on the mixture of \syndata + \traindata (y-axis) on two most tensive metrics gsm8k \textbf{(top)} and bbh-nocot-fs \textbf{(bottom)}. The performances includes different cases with 15K or 300K \traindata, masked or no-masked training.}.
    \label{fig:alone vs mixture}
\end{figure}

\begin{table*}[!ht]
\centering
\setlength{\belowcaptionskip}{-2pt}
\scalebox{0.6}{
\begin{tabular}{@{}l@{\hspace{0.5em}}c@{\hspace{0.5em}}c@{\hspace{0.5em}}c@{\hspace{0.5em}}c@{\hspace{0.5em}}c@{\hspace{0.5em}}c@{\hspace{0.5em}}c@{\hspace{0.5em}}c@{}}
\toprule
\textbf{Model} & \textbf{Size} & \textbf{mmlu} & \textbf{arc\_challenge} & \textbf{hellaswag} & \textbf{winogrande} & \textbf{truthfulqa\_mc2} & \textbf{agieval} & \textbf{avg} \\
 & & \textbf{(factuality)} & \textbf{(knowledge+ reasoning)} & \textbf{(knowledge+ reasoning)} & \textbf{(knowledge+ reasoning)} & \textbf{(truthfulness)} & \textbf{(instruct-follow)} & \\
\midrule
Baseline$_{4\text{epoch}}$ & 14.7k & 70.26 & 63.91 & 78.51 & 72.06 & 48.33 & 36.41 & 61.58 \\
Baseline$_{8\text{epoch}}$ & 14.7k & 70.38 & 63.53 & 79.58 & 70.6 & 49.29 & 36.31 & 61.61 \\
\midrule
NomaskedFiltered & 30.6k & 69.95 & 63.57 & 78.64 & 72.42 & 49.51 & 36.84 & 61.82 \\
MaskedFiltered & 25.7k  & 70.13 & 64.34 & 79.04 & 71.67 & 49.11 & 36.48 & 61.79 \\
\bottomrule
\end{tabular}
}
\caption{\small \small Performance comparison of different \syndata configurations and baselines with 15K TULU. \textsc{Nomasked\/Masked} indicates whether \syndata are trained with or without prompt masking. Easy to see that all those results are pretty close.}
\label{tab:performance_comparison_mc}
\end{table*}

Intuitively, it is easy to regard such OOD data as low-quality. However, in Table~\ref{fig:alone vs mixture}, we show that such a dataset alone can still be helpful and even achieve better results when compared to training with \syndata from no-prompt-masked alone. In fact, the performance degradation mainly occurs when mixing with \traindata. Thus, when measuring the "effectiveness" of \syndata, it is important to use the \traindata as reference. Moreover, this leave a future question that whether those generated \syndata is able to mix to other high quality data other than the original \traindata.

\section{More results on multi-choice metrics}

In Section~\ref{sec: mainresult}, we have shown the advantage of our methods on free-generation metrics. Nevertheless, we find that the proposed synthetic data generation methodology is less effective in multi-choice metrics.

\subsection{Details on evaluation metrics}
In multi-choice metrics, the learner are given a fixed set of candidates (e.g. A,B,C,D) and choose the result with maximum digits among those candidates. Here we consider the following metrics:

\paragraph{MMLU \citep{henderson-etal-2019-repository,hendrycks2021ethics}} (Knowledge)
It evaluates models across 57 diverse subjects, ranging from STEM fields to humanities and social sciences. This comprehensive test requires broad knowledge spanning elementary to professional-level expertise. Each task consists of multiple-choice questions, making it a robust measure of a model's acquired knowledge..

\paragraph{ARC Challenge \citep{Clark2018ThinkYH}} (Knowledge+reasoning)
It specifically focuses on grade-school science questions. The Challenge Set contains questions that cannot be answered by simple retrieval or word association methods, requiring both scientific knowledge and complex reasoning abilities. Questions often involve multi-step logical inference, causal reasoning, and the application of scientific principles to novel scenarios. 

\paragraph{hellaswag \citep{zellers2019hellaswag}} (Knowledge+reasoning) It is a challenging commonsense reasoning benchmark that consists of multiple-choice questions where systems must complete a sentence or short paragraph with the most contextually appropriate ending from four options. 

\paragraph{Winogrande \citep{sakaguchi2019winogrande}} (Knowledge+reasoning)
Winogrande is an evolved version of the Winograd Schema Challenge, designed to test common sense reasoning through pronoun resolution tasks. The dataset consists of sentences with ambiguous pronouns that can only be correctly resolved through understanding of context and real-world knowledge. What sets Winogrande apart is its carefully curated adversarial examples that minimize dataset artifacts, making it a more robust test of genuine reasoning capabilities. The questions require both implicit knowledge about how the world works and the ability to apply this knowledge in context-dependent ways.

\paragraph{TruthfulQA\_mc2 \citep{lin-etal-2022-truthfulqa}} (Truthfulness)
It is a specialized benchmark designed to evaluate a model's tendency to generate truthful versus false or misleading information. We have used its free-generation version in our main result. Here we instead use the multiple-choice version (mc2).

\paragraph{AGIEval \citep{zhong2023agieval}} (Instruct-follow)
AGIEval is a comprehensive benchmark designed to assess instruction-following capabilities and general intelligence in language models. It incorporates a diverse set of tasks that mirror real-world cognitive challenges, including professional certification questions, academic tests, and complex problem-solving scenarios. The benchmark is structured to evaluate not just the model's ability to understand instructions but also its capacity to apply knowledge in context-appropriate ways.

\label{app: more on 300K}
\begin{table*}[!ht]
\centering
\setlength{\belowcaptionskip}{-2pt}
\scalebox{0.7}{
\begin{tabular}{@{}l@{\hspace{0.5em}}c@{\hspace{0.5em}}c@{\hspace{0.5em}}c@{\hspace{0.5em}}c@{\hspace{0.5em}}c@{\hspace{0.5em}}c@{\hspace{0.5em}}c@{}}
\toprule
\textbf{Model} & \textbf{Size} & \textbf{TriviaQA} & \textbf{BBH-FS} & \textbf{BBH-COT-FS} & \textbf{GSM8K} & \textbf{TruthfulQA} & \textbf{Avg} \\
 & & \textbf{(Knowledge)} & \textbf{(Reasoning)} & \textbf{(Reasoning)} & \textbf{(Reasoning)} & \textbf{(Truthful)} & \\
\midrule
Baseline & 293.5k & 15.23 & 45.37 & 68.68 & 72.25 & 66.71 & 53.65 \\
\midrule
Nomask\_p09 & 322.0k & 14.51 & 39.93 & 64.97 & 72.48 & 66.59 & 51.70 \\
NomaskFiltered\_p09 & 309.5k & 13.39 & 46.84 & 65.07 & 71.95 & 67.56 & 52.96 \\
Nomask\_p07 & 321.0k & 15.29 & 41.81 & 65.46 & 73.92 & 67.81 & 52.86 \\
NomaskFiltered\_p07 & 309.1k & 14.43 & 39.87 & 66.43 & 74.00 & 66.22 & 52.19 \\
\midrule
masked\_p09 & 314.8k & 14.13 & 43.00 & 66.24 & 73.69 & 65.48 & 52.51 \\
maskedFiltered\_p09 & 306.8k & 14.95 & 43.25 & 67.76 & 73.62 & 65.61 & 53.04 \\
masked\_p07 & 313.8k & 15.75 & 41.87 & 65.93 & 73.01 & 65.85 & 52.48 \\
maskedFiltered\_p07 & 305.0k & 12.98 & 44.66 & 67.12 & 73.31 & 68.30 & 53.27 \\
\bottomrule
\end{tabular}
}
\caption{\small Performance comparison of different \syndata configurations with 300K TULU. Models are grouped by masking strategy (baseline, no mask, masked) and include filtered variants. The Size column shows the model size in thousands of parameters. Metrics evaluate knowledge, reasoning, and truthfulness capabilities. Each value represents the model's performance score on the respective benchmark.}
\label{tab:performance_comparison_full}
\end{table*}

\subsection{Results}
As shown in Table~\ref{tab:performance_comparison_mc}, in contrast to the significant improvements observed in free-generation metrics under 15K TULU, neither synthetic method demonstrates notable performance gains over the baseline. Furthermore, there is minimal difference in performance between prompt-masked and non-prompt-masked training approaches.

\section{More results on 300K parameters}

We present the comprehensive results in Table~\ref{tab:performance_comparison_full} using \traindata=300K TULU, including experiments with generation parameter $\text{top\_p} = 0.7$. Note that we excluded the $\text{top\_p} = 0.7$ configuration under the \traindata=15K TULU setting due to its inability to generate coherent sentences. The results demonstrate that all synthetic data generated using \traindata=300K TULU underperforms compared to the Baseline, with no significant variations across different $\text{top\_p}$ values. This observation reinforces our hypothesis that utilizing the full 300K dataset for \syndata generation yields outputs that closely mirror the original TULU distribution, regardless of other parameter choices.

\end{document}